# Uncertain Reasoning using Maximum Entropy Inference

*Daniel Hunter*

TRW/DSG
One Space Park, 02/1796
Redondo Beach, CA 90278

**Abstract**

The use of maximum entropy inference in reasoning with uncertain information is commonly justified by an information-theoretic argument. This paper discusses a possible objection to this information-theoretic justification and shows how it can be met. I then compare maximum entropy inference with certain other currently popular methods for uncertain reasoning. In making such a comparison, one must distinguish between static and dynamic theories of degrees of belief: a static theory concerns the consistency conditions for degrees of belief at a given time; whereas a dynamic theory concerns how one's degrees of belief should change in the light of new information. It is argued that maximum entropy is a dynamic theory and that a complete theory of uncertain reasoning can be gotten by combining maximum entropy inference with probability theory, which is a static theory. This total theory, I argue, is much better grounded than are other theories of uncertain reasoning.

## 1. Introduction

*Maximum relative entropy inference* (also known as *cross-entropy inference* or *minimum-information updating*) is a method for updating a probability distribution in the light of information in the form of new probabilities for propositions defined within the probability space. Where *prior* is the prior probability distribution over space $S = x_1, x_2, \ldots, x_n$ (I assume for the sake of simplicity that $S$ is finite) and $C$ is a set of constraints on the probabilities of subsets of $S$, maximum relative entropy inference tells us to choose as posterior that unique value of *post* that maximizes the relative entropy function

$$H(post, prior) = -\sum_{i=1}^{n} post(x_i) \log(post(x_i)/prior(x_i))$$

subject to the constraints given in $C$. The existence of this unique value of *post* is mathematically guaranteed provided $C$ is a consistent set of constraints.

The following sections explore the question of what justification can be given for thinking that maximum relative entropy inference is a reasonable way of updating probabilities. It is argued that the combination of probability theory and maximum relative entropy provides the best total approach to the problem of uncertain reasoning.

## 2. The Information-Theoretic Justification

The standard information-theoretic justification of maximum entropy inference goes as follows: when presented with information in the form of a set of probability constraints, we want to infer a probability distribution that satisfies those constraints but that, of all such distributions, is the least biased. To be least biased is to minimize the amount of information contained in the distribution (the distribution must have at least as much information content as the constraints since it satisfies them). Among the distributions satisfying the constraints there is a unique one, the maximum entropy distribution, that has minimum information. Hence maximum entropy inference selects the right distribution.

This argument has a great deal of intuitive force. However, we must be care-ful to be sure that its force does not depend upon an equivocation. The argument sounds plausible provided "information" is used in its ordinary sense: certainly when we make probabilistic inferences, we want to go beyond the given information to the smallest extent possible, but we cannot do this by minimizing just any function and calling the value of that function for a given probability space the "information" in that space. Thus we want assurance that maximizing entropy does come to the same thing as minimizing information in its ordinary sense.

Why should this even be an issue? It should be an issue because some leading information-theorists have denied that the standard mathematical (or engineering) definition of information captures its ordinary sense. For example, Hamming has written about the mathematical definition of information that:

> This is an engineering definition based on probabilities and is not a definition based on the meaning of symbols to the human receiver. The confusion at this point has been very great for outsiders who glance at information theory; they fail to grasp that this is a highly technical definition that captures *only part* [italicized in the original text] of the richness of the usual idea of information.
>
> To see how far this definition differs from "common sense" consider the question : "What book contains the most information?" We, of course, have to standardize the question by stating the book, page, and type-font sizes, as well as the variety (alphabet) of type fonts to be used. Once this is done, the answer is clearly : "The book with the most information is the one with the type chosen completely and uniformly at random!" Each new symbol will come as a complete surprise. ... [4] p. 103.

Abramson, in [1] p.2, makes a similar claim. In talking about Shannon's 1948 paper "A Mathematical Theory of Communication", which initiated the field of information theory, he states : "In terms of the colloquial meaning of information, Shannon's paper deals with the carriers of information--symbols--and not with information itself. It deals with communication and the means of communication rather than that elusive end product of communication--information."

Both these quotation from experts on information theory suggest a divergence between the mathematical sense of "information" and its ordinary or colloquial sense. However, their reasons for thinking such a divergence exists are confused. To bring out what the confusion is, let me focus on Hamming's example of the information content of a book. His claim is that a book whose symbols occur randomly contains the maximum amount of information in the mathematical sense but such a book contains mere gibberish and so conveys no information in the ordinary sense.

This example is highly misleading. We have to keep in mind that information is defined *relative a probability space*. And now we must ask, "Relative to what space is the information of the book maximum?" The answer must be "Relative to the space whose points are occurrences of certain symbols (i.e.the symbols that make up the text), where the points are equiprobable." Call this space B. Indeed, H(B) is maximal when the members of B are equiprobable. However, in reading a text, we are rarely concerned with the information it provides us concerning the occurrence of symbols; rather, we are concerned with what it tells us about certain non-linguistic states. By virtue of linguistic conventions, there is a correlation between linguistic entities and states of the world. If the linguistic entities are declarative sentences and we have reason to believe that those sentences are true, then the occurrence of certain symbols will give us evidence as to the existence of certain states of the world. Let W be a probability space whose events are occurrences of certain non-linguistic events. Then we are in general concerned not about B in itself but about the information B imparts about W. The information that B imparts about W is written "I(W;S)"--i.e. it is the increase in information about W that we get on learning which events in B obtain. Now the initial uncertainty concerning W is H(W), the entropy of W. The uncertainty that remains when the truth about B is learned is H(W | B), the conditional entropy of W with respect to B, which is defined by:

(1) $H(W|B) = \sum prob(w,b) log(1/prob(w|b))$,

where $prob(w,b)$ is the joint probability of $w$ and $b$, $prob(w|b)$ is the conditional probability of $w$ with respect to $b$, and the sum is over $w$ in W and $b$ in B. Hence the increase in information that occurs on learning which element of B obtains is given by:

(2) $I(W;B) = H(W) - H(W|B)$.

Substituting B for W in (2), we see that the information gained about B itself when the truth about B is learned given by:

(3) $I(B;B) = H(B)$.

Normally, when we say that, for example, one book contains more information than another, we are speaking about conditional information, information that the books give us about some space other than the space of occurrences of symbols in the books. So if W is the space of interest and B the space of symbol occurrences in the book, for an informative book I(W;B) has a high value. When Hamming speaks of the book whose symbols occur randomly as containing the most information, he is speaking of I(B;B), not I(W;B). But this is not a different sense of "information"; it is only a difference in the space about which we want information. Furthermore, maximizing I(B;B) may well be incompatible with maximizing I(W;B). Suppose that B is random in a strong sense: that each member of B is equiprobable and moreover that the probability of any member of B given an event outside of B is equal to the probability of that member (i.e. there is nothing outside of B that would help us predict which member of B will occur). Then I(B;B) will be maximal but I(W;B) will be zero.

Thus Hamming and Abramson have given no good reason for believing that the mathematical sense of "information" differs from its ordinary sense. What is more, a plausible argument can be given that the mathematical definition does indeed precisely capture the ordinary notion of "information": There is a well-known mathematical result that the logarithm is the only function f that satisfies the condition that f(xy) = f(x) + f(y) (together with some other trivial conditions such as that f is continuous over its domain). But our intuitive notion of amount of information satisfies this condition. For if events E1 and E2 are independent, then the amount of information gained from learning that both occur should be the sum of the amounts of information gained from learning of the occurrence of each separately. If we assume that the amount of information gained is a function of the probability of the event, then the condition becomes I(Prob(E1)Prob(E2)) = I(Prob(E1)) + I(Prob(E2)), where I denotes the information of a given probability. Hence I(x) can be defined as -log(x) (since x is in the range (0,1], we need to take the negative of the log so that I(x) will be a positive quantity). Then to get the entropy of an entire probability distribution, we average over the entropy of each element and so arrive at the standard mathematical definition of entropy.

## 3. The Axiomatic Justification

The intuitive argument for maximum entropy inference still goes through, then: maximum entropy inference really does minimize information in its ordinary sense. Some might argue, however, that this intuitive argument is not needed, since the validity of maximum entropy inference has been axiomatically derived by Shore and Johnson in [6]. It is worth spending some time on this derivation, since it raises interesting epistemological issues. Shore and Johnson's result is a result of the type, "The unique function satisfying conditions $C_1,...,C_n$ is function F." Results of this type can be important since they reduce the question of whether or not a specified mathematical function correctly represents a given intuitive concept to the question of whether or not the concept satisfies the stated conditions (We showed the use of such a result earlier in the argument that -log(pr(e)) gives the information in event e). However, it is not always easy to tell, in a non-circular fashion, whether or not the given concept satisfies the stated conditions. Shore and Johnson state four axioms that they take to characterize an information operator that takes a prior distribution and a

23

set of constraints the posterior distribution is to satisfy and yields a particular posterior distribution satisfying those constraints. Intuitively, we want this information operator to yield the best estimate of the posterior distribution in the light of the prior distribution and the information about the posterior. Instead of giving an explicit definition of this operator, Shore and Johnson characterize it axiomatically and prove that any interpretation of it that satisfies the axioms must yield the same result as maximum relative entropy.

This is a valuable result, for the reasons mentioned, but it is not clear that it should increase our confidence that maximum relative inference is the correct technique for determining a posterior. For not all the axioms that Shore and Johnson state are clearly axioms that should hold of such an information operator. Consider, for example, their axiom four. In this axiom $S_1,...,S_n$ is a partition on space D; for each i, $1 \leq i \leq n$, $I_i$ is information about the conditional density within $S_i$; $I$ is the conjunction of the $I_i$; $prob * Si$ is the relativization of probability distribution $prob$ to set $Si$; $M$ gives the probability of being in each of the $S_i$; and $post(p,i)$ denotes the result of applying the information operator to probability distribution $p$ with respect to information $i$ about the posterior. Axiom four reads:

(4)  $post(prior, I\&M) * Si = post(prior * Si, Ii)$.

Consider one consequence of (4) when $I$ is empty. If $I$ is empty, then $post(prior, I\&M) * Si = post(prior, M) * Si$ and $post(prior * Si, Ii) = prior * Si$. Hence in the special case in which $I$ is empty, we have this version of (4):

(4a)  $post(prior, M) * Si = prior * Si$.

If A is an arbitrary subset of D, (4a) can be more perspicuously written as:

(4b)  $post(prior, M)(A | Si) = prior(A | Si)$.

(4b) says that new information about the probability of members of the partition does not affect conditional probabilities within members of the partition. Is this a reasonable condition to impose on the information operator? Skyrms [8] has argued that there are cases in which (4b) fails, the particulars of the case depending upon what concept of probability is in question. Suppose, to use an example from Skyrms, that constraint M gives the physical probability (propensity) of members of the partition S. Then M may give us information about events in D not available from knowing which member of S obtains. For example, if we know that the physical probability of a coin's coming up heads on the next flip is 2/3, then this gives us information about the coin that we wouldn't have simply from knowing that the next flip came up heads -- e.g, we have reason to believe it did not come from the U.S. mint. Thus if we let M = "The physical probability of heads on the next flip is 2/3", H = "The coin lands heads on the next flip", and A = "The coin came from the U.S. mint", then (4b) fails: post(prior,M)(A | H) is less than prior(A | H).

Consider another case, one in which probability is personal probability. If A is a proposition about an action you will perform in the future and the members of S are propositions about the world relevant to deciding whether or not to perform the action, then (4b) would seem to fail. For the probability of your performing an action would depend more strongly on your personal probabilities for the states of the world than on the states of the world themselves. As an example, consider the decision-theoretic problem of the lady and the tiger: you are to choose one of two doors; behind one door is a tiger, behind the other is a beautiful princess. If you choose the door behind which there is the princess, she is yours; if the door behind which there is a tiger, you will be eaten alive. Label the doors "door 1" and "door 2" and let T be the proposition that there is a tiger behind door 1. If we let D1 be the action of choosing door 1, then the probability of D1 depends more upon the chooser's personal probability for T than it does upon the truth of T. Thus if the probability of T is initially 0.5 and the new constraint C is that the probability of T is 0.8 (perhaps the chooser thinks he hears ominous growling behind door 1 -- the growling is probably from the tiger although it's possible the princess ate something for lunch that didn't agree with her). Then post(prior,C)(D1 | T) should be lower than prior(D1 | T), unless the chooser believes at

the time at which he has the prior that the truth of T will lead, before the moment of decision, to a personal probability for T greater than or equal to 0.8 -- but he need not believe this; he may expect no new information about what is behind the doors to be forthcoming.

The above examples cast doubt on Shore and Johnson's axiom four. I believe that the supporter of maximum entropy can make a plausible response to these examples, but space does not permit plunging into this complicated debate. In any case, the moral I wish to draw from this debate is that an axiomatic justification of the sort Shore and Johnson give is convincing only to the degree that the individual axioms are convincing. Since Shore and Johnson's axiom four is not obviously true, they cannot claim to have *proven* the validity of maximum relative entropy inference. Rather, I think the best argument we have now for maximum relative entropy inference is the intuitive information-theoretic one. If we think this argument sound, then we should also think that axiom four is true (since maximum relative entropy inference does satisfy axiom four), but then to use axioms one through four to argue for maximum relative entropy would be circular.

### 4. Other Methods of Reasoning with Uncertainty

I now want to turn my attention to the question of how maximum entropy inference compares with other currently popular methods of uncertain inference. One thing that must be kept in mind in making such a comparison is that theories of uncertain inference generally have two components, one a view concerning the conditions under which a belief system at a particular time is consistent and another a view about how a given belief system should be changed in the light of new evidence. Call the first component the "static component" and the second the "dynamic component". To illustrate the difference between these two components, consider the following example: I now believe a given die is fair; I therefore assign probability 1/6 to each of the possibilities of a certain number coming up on a throw of the die; what probability should I *now* assign to the disjunction "Either one comes up or two comes up"? This is a problem addressed by the static component of a theory since it concerns the consistency of a belief system at a single time. Suppose, however, that I learn that the die is not fair and that in fact the expected value of the number showing on the die is 4.5. Then the problem of how to *revise* my system of belief to accommodate this new information, is a problem addressed by the dynamic component of the theory since it concerns how my belief system changes through time. The static component of the theory can give no answer to the dynamic problem : there will in general be many consistent belief systems compatible with the new information.

Much confusion can result if the above distinction is ignored. In comparing two theories of reasoning with uncertainty, one must compare each component of one with the corresponding component of the other. To compare the static component of one theory with the dynamic component of the other only invites confusion. It seems to me that some commentators have fallen into this trap of comparing apples and oranges, especially when one of the theories in question is maximum entropy. The reason for this is that these commentators have not seen clearly that maximum entropy is a dynamic, rather than a static, theory. The dynamic view of maximum entropy is common in the philosophical literature on the subject (e.g, see [8], [9] and the references therein), but since it is apt to be controversial elsewhere, let me spend some time defending this view of maximum entropy.

On this view of maximum entropy inference, maximum relative entropy, or cross-entropy, is the basic notion, and what is known simply as "maximum entropy" is maximum relative entropy with respect to a uniform prior. Thus we always think of maximum relative inference as a way of passing from one probability distribution to another. In this sense maximum entropy inference addresses only the dynamic problem and not the static one. When maximum entropy inference is applied to a case in which there are no non-trivial constraints, one might think that it is being used to get a prior distribution and therefore to solve a static problem. However, if we think of maximum entropy as maximum relative entropy with respect to a uniform prior, then we are already assuming a uniform prior, which, of course, remains uniform in the absence of any new constraints; regarding maximum *relative* entropy as the basic notion is justified by the fact that in the continuous case, there is no straightforward way of generalizing the nonrelative maximum entropy formula for the discrete case; as Jaynes points out ([5], pp.59-60 and p. 124) the best that can be done is to change the summation to an integral at the same time introducing an invariant measure m(x) over the space in question. This measure m(x) represents a distribution of total indifference--i.e. a distribution expressing complete ignorance; and maximum entropy does not provide a method for determining this indifference measure.

Some of the putative rivals of maximum entropy inference do not challenge it directly because they primarily address the static problem. The max-min rules for combining certainty factors in the inference system of MYCIN [7] or for combining possibilities in fuzzy logic [10], for example, are really rivals of standard probability theory, not of maximum entropy. To the extent that these nonstandard theories address the dynamic component, the divergence from maximum entropy is often quite small. In the inference system of MYCIN, for example, when the evidence is known with certainty, updating reduces to conditionalization on the evidence; and when the evidence is not certain the certainty factor of the hypothesis on the basis of the evidence is multiplied by the certainty factor of the evidence to yield the final degree of belief in the hypothesis, which gives a rough approximation to the result of maximum entropy inference (the approximation is better the more certain the evidence is). What is distinctive about the approach to uncertainty implemented by the authors of MYCIN is not the updating, but the static rules for combining certainty factors.

However, advocates of maximum entropy inference generally view it as part of a package whose static component is standard probability theory. And it is the entire package that should be evaluated. Peter Cheeseman [2] has done a much needed job in defending probability theory against current objections. Let me say some words in its defense also. Attempts to replace probability theory wrongly ignore the coherence arguments for the axioms of probability theory given by de Finetti and others (e.g., see [3] and [8]). These arguments show that anyone whose degrees of belief violate the probability axioms could have a "Dutch book" made against him --i.e. he could be presented with a series bets that he would have to regard as fair in light of his beliefs, but he would be guaranteed to lose money on the bets. Now the formulation in terms of bets is unnecessary. The formulation by de Finetti ([3] pp.13-14) may be more agreeable to those averse to thinking in terms of gambles: the idea is to view a decision about the occurrence of events $E_1, E_2, ..., E_n$ as the specification of numbers $x_1, x_2, ..., x_n$ such that the loss due to choosing those numbers is

$$L = (v(E_1) - x_1)^2 + (v(E_2) - x_2)^2 + \cdots + (v(E_n) - x_n)^2$$

where $v(E_i) = 1$ if $E_i$ occurs and $= 0$ otherwise; then define a decision D to be *admissible* if there does not exist a decision whose loss is *always* less than that of the D (i.e. no matter which events occur). de Finetti shows that if the x's are chosen so that the decision is admissible then they have to satisfy the axioms of probability -- they must play the role of probabilities in the reasoning of the agent.

This reduction of probability theory to decision theory is of great importance. Not only does it justify the axioms of probability, it also sheds light on the nature of probabilities as used in decision problems. For the reduction to make sense, probabilities must express an agent's uncertainty about the facts; hence the probabilities must be relative to the agent in some sense (but not necessarily subjective-- as Jaynes argues [5], such probabilities may be objective in that two agents with the same state of knowledge should agree on probabilities). Moreover, there has to be a definite set of facts about which the agent is uncertain for the loss function to be well-defined. An agent's uncertainty about the truth of a proposition must be distinguished from the proposition itself being inherently vague. For example, it is one thing to be unsure about whether or not Jones is tall because one is ignorant of Jones' height but quite another to be unsure of this because Jones is a borderline case of a tall man. The uncertainty, or better, vagueness, attendant to the second case may perhaps be appropriately analyzed by some version of fuzzy logic. But fuzzy logic is not appropriate in cases of the first sort. Zadeh's view seems to be that fuzzy logic is appropriate to cases in which the uncertainty is non- statistical in nature and does not involve the "the notion of repeated experimentation." ([10] p. 152). If this means that the uncertainties cannot be given a frequency interpretation, then this is incorrect, for there are decision problems involving uncertainties that cannot plausibly be given a frequency interpretation, but which are really uncertainties about the truth of non-vague propositions.

## 5. Conclusion

Fuzzy logic and other rivals of probability theory face a formidable challenge from de Finetti's decision-theoretic foundation for probability theory. Moreover, since these non-standard theories are primarily static theories, they do not directly challenge maximum relative entropy inference. Hence in view of the strong information-theoretic justification for maximum relative entropy inference, its combination with probability theory appears to be the best approach to reasoning with uncertainty.